\documentclass[10pt,journal,compsoc]{IEEEtran}
\ifCLASSOPTIONcompsoc
  \usepackage[nocompress]{cite}
\else
  \usepackage{cite}
\fi
\ifCLASSINFOpdf
\else
\fi

\usepackage{amsmath,amsfonts}
\usepackage{algorithmic}
\usepackage{algorithm}
\usepackage{array}
\usepackage[caption=false,font=normalsize,labelfont=sf,textfont=sf]{subfig}
\usepackage{textcomp}
\usepackage{stfloats}
\usepackage{url}
\usepackage{verbatim}
\usepackage{graphicx}
\usepackage{cite}
\usepackage{multirow}
\usepackage{color}
\usepackage[table]{xcolor}
\usepackage{arydshln} 
\usepackage{amsmath,amsfonts}

\usepackage[caption=false,font=normalsize,labelfont=sf,textfont=sf]{subfig}
\usepackage{textcomp}
\usepackage{stfloats}
\usepackage{url}
\usepackage{verbatim}
\usepackage{graphicx}
\usepackage{cite}
\usepackage{booktabs}
\usepackage{arydshln}
\usepackage{url} 
\usepackage[colorlinks=true, linkcolor=blue, citecolor=green, urlcolor=cyan]{hyperref}

\definecolor{lzh}{RGB}{153,153,255}

\begin{document}
%
\title{Visual Text Meets Low-level Vision: A Comprehensive Survey on Visual Text Processing}

\author{Yan Shu, Weichao Zeng, Zhenhang Li, Fangmin Zhao, Yu Zhou

 \thanks{
    • Y. Shu is with the Institute of Information Engineering, Chinese Academy of Sciences, China (e-mail: shuyan9812@gmail.com).
  }

\thanks{  • W. Zeng, Z. Li, F. Zhao and Y. Zhou are with the Institute of Information Engineering, Chinese Academy of Sciences, China, and also with the School of Cyber Security, University of Chinese Academy of Sciences, China (e-mail: zengweichao@iie.ac.cn; lizhenhang@iie.ac.cn; zhaofangmin@iie.ac.cn; zhouyu@iie.ac.cn).}

\thanks{ •  Corresponding author: Y. Zhou.}
}

\IEEEtitleabstractindextext{%
\begin{abstract}
Visual text, a pivotal element in both document and scene images, speaks volumes and attracts significant attention in the computer vision domain. Beyond visual text detection and recognition, the field of visual text processing has experienced a surge in research, driven by the advent of fundamental generative models. However, challenges persist due to the unique properties and features that distinguish text from general objects. Effectively leveraging these unique textual characteristics is crucial in visual text processing, as observed in our study. In this survey, we present a comprehensive, multi-perspective analysis of recent advancements in this field. Initially, we introduce a hierarchical taxonomy encompassing areas ranging from text image enhancement and restoration to text image manipulation, followed by different learning paradigms. Subsequently, we conduct an in-depth discussion of how specific textual features—such as structure, stroke, semantics, style, and spatial context—are seamlessly integrated into various tasks. Furthermore, we explore available public datasets and benchmark the reviewed methods on several widely-used datasets. Finally, we identify principal challenges and potential avenues for future research. Our aim is to establish this survey as a fundamental resource, fostering continued exploration and innovation in the dynamic area of visual text processing.  A project associated with this survey is available at \url{https://github.com/shuyansy/Survey-of-Visual-Text-Processing}.
\end{abstract}

}

\maketitle

\IEEEdisplaynontitleabstractindextext

%
\IEEEpeerreviewmaketitle

\IEEEraisesectionheading{\section{Introduction}\label{sec:introduction}}

\begin{figure*}[t]
 \setlength{\abovecaptionskip}{0cm} 
\begin{center}
\includegraphics[width=0.9\textwidth]{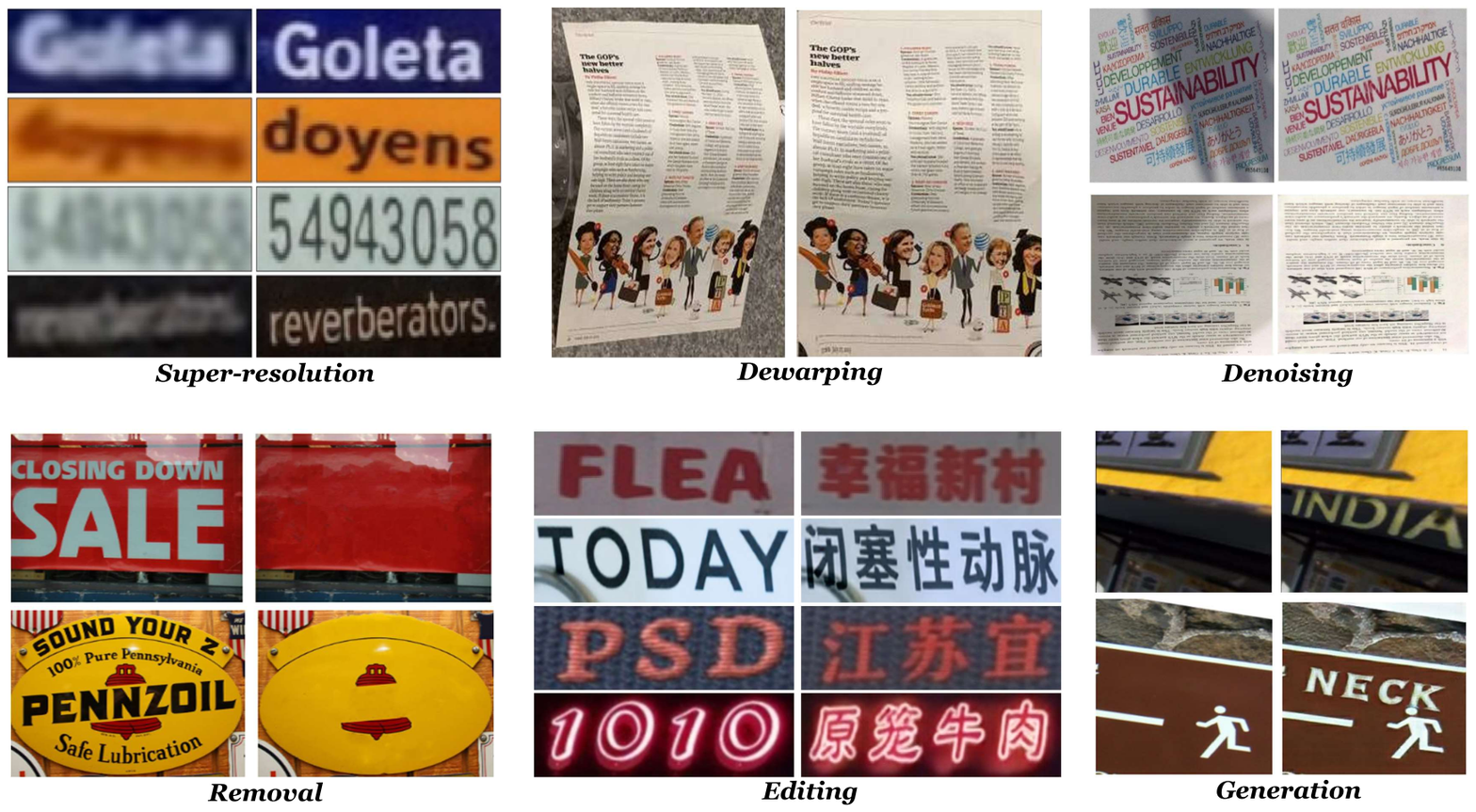}
\hfill
\end{center}
\vspace{-40pt}
\caption{ Visualization samples of visual text processing tasks. The top row is the text image enhancement/restoration, including super-resolution \cite{noguchi2024scene}, dewarping\cite{Ma2018DocUNetDI}, and denoising \cite{Lin2020BEDSRNetAD}. The bottom row is text image manipulation, including text removal \cite{peng2023viteraser}, text editing \cite{yang2023self}, and text generation \cite{zhan2019spatial}.        
}
\label{fig1}
\end{figure*}

\IEEEPARstart{V}{isual} text, which means the embedded text element in images, plays an important role in image/video retrieval \cite{fang2023uatvr}, assistance for visually impaired people, scene understanding, document artificial intelligence, etc. According to the text image types, visual text can be categorized into document text and scene text, which are embedded in document images and scene images respectively. The domain of visual text research bifurcates into two primary branches: text spotting and text processing. There has been a substantial increase in the body of work focusing on text spotting in the wild. This research evolution is mapped from the era preceding deep learning to the current paradigm dominated by deep learning techniques, which is a progression underscored by numerous studies \cite{east,crnn,db,masktext}. Thorough surveys have been documented in the literature \cite{ye2014text,zhu2016scene,yin2016text,liu2019scene,lin2020review,khan2021deep,chen2021text}, encapsulating these developments.

Reviews by Ye et al. \cite{ye2014text} and Zhu et al. \cite{zhu2016scene} are centered on image-based text detection and recognition works mostly utilizing traditional handcrafted features. Reviews by Liu et al. \cite{liu2019scene} and Lin et al. \cite{lin2020review} have shifted the emphasis towards deep learning frameworks for detecting and recognizing scene text. Furthermore, Chen et al. \cite{chen2021text} delve comprehensively into scene text recognition technologies. Despite these scholarly contributions on text spotting (including detection and recognition), the literature still lacks a unified survey that integrates the full gamut of visual text processing research.

The domain of visual text processing includes two main categories: text image enhancement/restoration and text image manipulation, as demonstrated in Figure \ref{fig1}. The enhancement/restoration category includes: (i) Text image super-resolution, which enhances the resolution and clarity of text within low-resolution image; (ii) Document image dewarping, which corrects geometric distortions pivotal for digitization workflows; (iii) Text image denoising, which aims at reducing noise and improving image quality. In contrast, the manipulation category comprises: (i) Text removal, which eliminates text from image and restores pixels of underlying backgrounds; (ii) Text editing, which alters text content while preserves its original aesthetic; (iii) Text generation, which synthesizes text image with diverse appearances that maintain visual authenticity. Other related topics include text segmentation and editing detection. Visual text processing is crucial in numerous practical applications. Text image enhancement and restoration tasks primarily focus on augmenting the quality of low-fidelity images. This includes correcting text positioning through dewarping, and enhancing readability via super-resolution or denoising, crucial for boosting text recognition and understanding accuracy \cite{crnn,ASTER,kil2023prestu}. Meanwhile, text image manipulation techniques play a vital role in privacy protection \cite{inai2014selective} through text removal, image translation \cite{fragoso2011translatar} via editing, and enhancing augmented reality interfaces \cite{abu2018augmented} through text generation.

Visual text processing is a subfield of low-level computer vision, but is more specifically focused on text pixels. Moreover, from the methodological point of view, it is tightly related to generative artificial intelligence (AI). In the realm of generative AI, the field has experienced significant advancements owing to the evolution of deep learning, particularly marked by the development of groundbreaking frameworks like Generative Adversarial Networks  \cite{goodfellow2014generative} and diffusion models \cite{ho2020denoising,croitoru2023diffusion}. On one hand, these general paradigms endow visual text processing methods with robust capabilities due to the inherent similarities between texts and general objects. On the other hand, they also encounter numerous challenges, as texts possess distinct characteristics that set them apart from general objects. For example, scene text instances may vary in languages, colors, fonts, sizes, orientations, and shapes.

To address these challenges, researchers have investigated a range of text-related features, encompassing structure (layout and orientation), stroke (character glyph), semantics (language information), style (color and font), and spatial contexts (background texture and depth). The widespread use of text-related tasks, employing either fully annotated data under strong supervision or designing weak supervision methods, facilitates the extraction of specific text features. Furthermore, the burgeoning fields of multi-task architectures \cite{zhang2021survey} and conditional generative models \cite{mirza2014conditional,chrysos2018robust} allow for the flexible integration of various text features into different visual text processing frameworks, resulting in notable enhancements.

In this survey, we provide a comprehensive, multi-perspective overview of the most recent advancements in deep-learning-based visual text processing works. Initially, we classify existing works according to the purpose of processing, followed by different learning paradigms, thereby establish a hierarchical taxonomy. Subsequently, we engage in in-depth discussions of seminal works within various text feature categories, focusing particularly on the seamless integration of text characteristics and network designs. Following this, we detail benchmark datasets, evaluation metrics, and corresponding experimental result comparison. Finally, we highlight the current research challenges and suggest potential directions for future investigations. 

In summary, our contributions are as follows:

(i) Despite the existence of numerous surveys on text detection and recognition, this is the first work to offer a comprehensive literature review specifically focused on visual text processing works.

(ii) We have developed a multi-perspective categorization scheme for visual text processing works. This not only entails a hierarchical taxonomy based on different tasks and learning paradigms but also delves deeply into various distinct text features.

(iii) We present a thorough overview of various datasets from different text processing tasks, along with a critical assessment of the performance of contemporary works.

(iv) We identify and summarize the open challenges in current research, offering our insights on promising directions for future exploration in this field.

\begin{figure*}[h]
 \setlength{\abovecaptionskip}{0cm} 
\begin{center}
\includegraphics[width=1\textwidth]{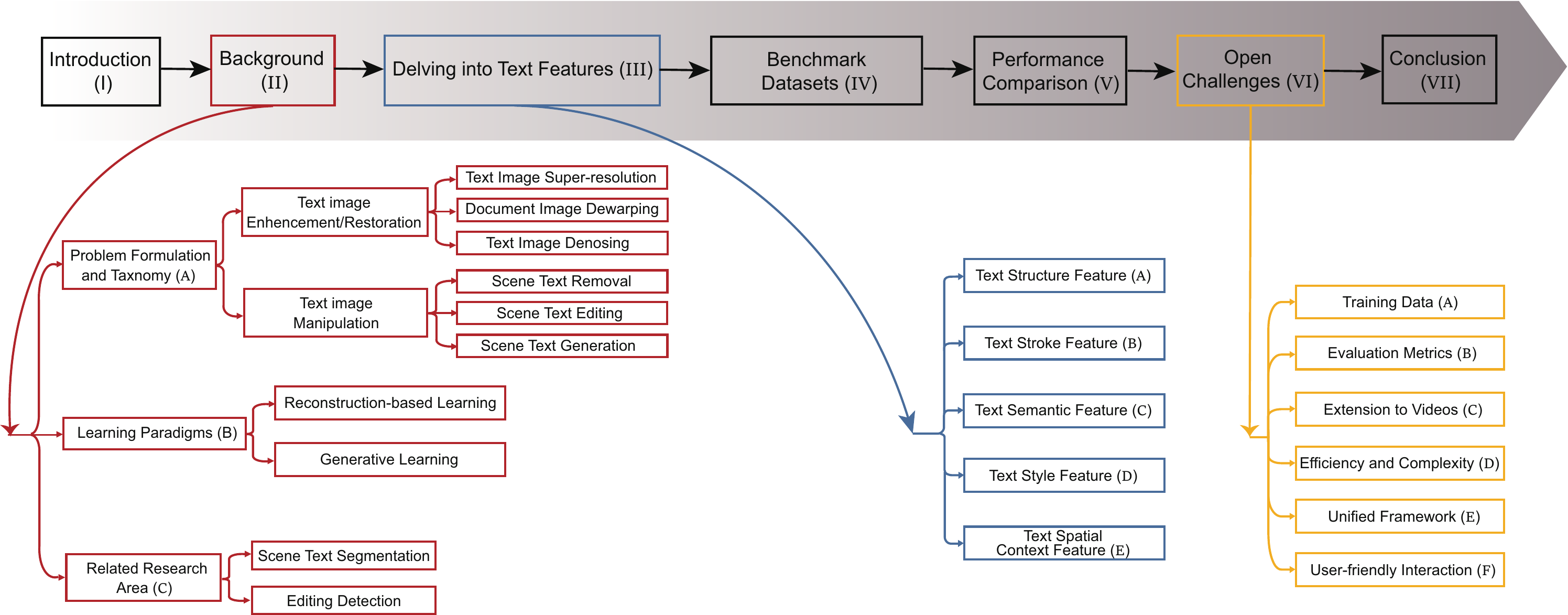}
\hfill
\end{center}
\caption{Main structure of this survey.  Initially, we introduce a hierarchical taxonomy  from image enhancement and restoration to image manipulation, followed by different learning paradigms. Subsequently, we
conduct an in-depth discussion of how specific textual features are
integrated into various tasks. Furthermore, we explore public datasets and benchmark the reviewed methods. Finally, we identify open challenges  for future research.} 
\label{fig2}
\end{figure*}

The organization of this survey is illustrated in Figure \ref{fig2}. Section \ref{section2} provides a concise background on problem-related taxonomy and related research areas. Section \ref{section3} thoroughly reviews representative works in this field, emphasizing their seamless integration with specific text characteristics. Section \ref{section4} examines the available datasets. Section \ref{section5} compares the reviewed works on benchmarks. Section \ref{section6} discusses the existing open challenges in the field and offers insights into potential future developments. Section \ref{section7} concludes this survey.

\section{Background}
\label{section2}

\subsection{Problem Formulation and Taxonomy}

Formally, let  $\boldsymbol{X}$ and $\boldsymbol{Y}$ denote the input and output spaces, respectively. Solutions in deep learning-based visual text processing typically aim to learn an optimal mapping function which can be mathematically represented as $f^*:\boldsymbol{X}\rightarrow\boldsymbol{Y}$. Differentiated by the nature of  $\boldsymbol{Y}$, we categorize existing works into two main areas: text image enhancement/restoration and text image manipulation, wherein each area consists of various tasks characterized by specific concerns. In the following subsection, we illustrate the specific context of $\boldsymbol{X}$ and $\boldsymbol{Y}$ in each area and task respectively.

\subsubsection{Text Image Enhancement/Restoration}

Text images captured in natural scenes or documents often suffer from low fidelity due to factors such as low resolution, distortion, and noise interference. To address this issue, various methods have been proposed aiming to enhance or restore the quality of text images. These methods can be further categorized into super-resolution, dewarping, and denoising. In this context, $\boldsymbol{Y}$ should maintain semantic consistency with $\boldsymbol{X}$, while the pixel-space distribution should be refined to align with the standards of human evaluation.

\textbf{Text Image Super-resolution} Text image super-resolution (SR)  aims to reconstruct high-resolution (HR) text images $\boldsymbol{Y}$ from their low-resolution (LR) counterparts $\boldsymbol{X}$, which suffer from diverse degradations. This task facilitates subsequent text recognition task \cite{qiao2020seed}  drastically. While sharing commonalities with the broader scope of general image super-resolution, text image SR presents unique challenges. Primarily, it is a foreground-centric task where the quality of the foreground text is paramount in evaluation, overshadowing background texture restoration. Moreover, successful restoration must preserve the textural continuity and, crucially, the semantic integrity of the text across HR and LR versions. This is particularly critical for languages with complex character structures, where minor stroke discrepancies can significantly alter visual perception and lead to misinterpretation. Furthermore, the variability of degradative factors in real-world scenarios (such as equipment quality, lighting conditions, and compression algorithms) poses additional obstacles to the generalizability of the proposed methods.


\textbf{Document Image Dewarping} Document image dewarping (DID) is expected to convert distorted document images into flat images based on coordinate mapping. Uncontrollable factors like suboptimal camera angles, positions, and document physical deformations significantly impede the visual interpretation of document images, detrimentally affecting subsequent processes including text recognition \cite{qiao2021pimnet}, table structure recognition \cite{shen2023divide}, and visual information extraction. In this case, $\boldsymbol{X}$ refers to a distorted document image as input, while $\boldsymbol{Y}$ is the coordinate mapping between the source image and predicted flatten image. Recently, DID has emerged as a critical research domain. Despite significant advancements, DID continues to face substantial challenges. Current methods often depend on predefined constraints, which can lead to mode collapse in diverse application scenarios, such as those involving documents with unclear or incomplete outlines. Furthermore, while existing DID techniques generally require highly accurate ground truth for effective outcomes, the existing well annotated datasets are all synthetic and  the vast reservoir of unlabeled real-world data remains underutilized.

\textbf{Text Image Denoising} 
Text image denoising (TID) is dedicated to reducing the negative effects, such as shadows, stains, and watermarks, on the captured text image $\boldsymbol{X}$, aiming for a noise-free prediction $\boldsymbol{Y}$. This enhancement improves readability and the performance of downstream applications like text detection \cite{shu2023perceiving}. Unlike denoising natural images, TID requires a meticulous approach to maintain the integrity of text structure and content. Reflecting the diversity of noise types, research in this domain is generally divided into two primary categories: illumination removal, addressing issues like underexposure, overexposure, and shadows; and impurity removal, a critical aspect of TID, which concentrates on eradicating fragmented noise such as ink artifacts \cite{Huang2019DeepEraseWS}, watermarks \cite{Yang2023DocDiffDE}, and stamps \cite{Yang2023MaskGuidedSE}. 
It remains an open question whether a single, large model can effectively eliminate various types of noise, provided that the training data are sufficient.




\subsubsection{Text Image Manipulation}



Texts within images in natural scenes often require manipulation to serve various objectives, such as privacy protection, image translation, and Augmented Reality (AR)-related applications. Existing works mainly focus on text image removal, text image editing and text image generation/synthesis. For output image $\boldsymbol{Y}$ in this context, the visual outcomes should either maintain consistency with input image $\boldsymbol{X}$ or faithfully comply with input condition $\boldsymbol{X}$, while the text content should be either eliminated, modified or appended. 

\textbf{Scene Text Removal}
Scene text removal (STR) represents an essential process that entails the deletion of text from natural images and the seamless infilling of these areas with contextually appropriate background pixels. In this case, $\boldsymbol{Y}$ is a text-free background image. Given the prevalence of text in images, especially on social media, STR has become critical for privacy protection. This task bifurcates into two essential sub-tasks: text localization to identify textual region and background reconstruction to replace the text. Recent progress in STR methodologies has led to the development of two primary approaches: direct removal which only takes $\boldsymbol{X}$ as input and auxiliary removal takes $\boldsymbol{X,M}$ as input, where $\boldsymbol{M}$ is the binary text region segmentation mask. Compared to direct removal methods, auxiliary removal methods typically demonstrate superior results owing to their precise detection indicators.

\textbf{Scene Text Editing} Scene text editing focuses on attribute changes, style transfer, and content modifications.
The overarching goal is to  replace text in a way that seamlessly integrates with the background, thereby minimizing any disruption to the image's overall appearance. Text editing tasks, though varied in approach, typically involve three core subtasks: text alteration, background restoration, and image integration. Recent advancements in this field have led to the categorization of methods into two principal groups: style editing and content editing.  In style editing, $\boldsymbol{Y}$ maintains the same content as $\boldsymbol{X}$, but with alterations to the appearance, color, and background of characters. Notable advancements in style editing techniques \cite{nakamura2019scene,gomez2019selective} have enhanced image text processing tools, leading to smarter, more automated applications across various domains. Conversely, content editing, which is distinct from style editing's aesthetic focus, aims for $\boldsymbol{Y}$ to preserve the original textual style of $\boldsymbol{X}$ while changing its words or characters. This branch of research typically focuses on two tasks: adapting target text to match the style of a given image and integrating it seamlessly into the original text area.

\textbf{Scene Text Generation} In the deep learning era, the scale of the training dataset crucially influences model performance. The task of scene text detection and recognition, however, requires extensive manual effort in annotating text boxes and corresponding contents, leading to a limited availability of real-world data. To overcome this data scarcity, text image synthesis methods have emerged, providing an alternative to the labor-intensive process of manual annotation for scene text datasets. Despite this, the quality of synthetic images remains a critical issue, which should resemble the distribution of real data.

\subsection{Learning Paradigms}

In this section, we examine the prevalent learning paradigms corresponding to distinct mapping functions in different tasks. 


\subsubsection{Reconstruction-based Learning}
Reconstruction-based learning methods are essential for restoring and enhancing text images, aiming to correct distortion, enhance clarity, and improve overall readability. To this end, pixel amplification methods, coordinate registration methods, and segmentation methods have emerged respectively.

\textbf{Pixel Amplification Methods}
The core of these methods involves using a loss function, typically Mean Squared Error (MSE), to minimize the pixel-wise differences between the enhanced output and the ground truth. Architectures like U-Net \cite{Ronneberger2015UNetCN} and Vision Transformers \cite{dosovitskiy2020image} are often employed.

Predominantly, text image super-resolution methods, which are modeled as a preprocessing step for OCR systems, utilize MSE Loss calculated between between high-resolution (HR) images and reconstruction results of low-resolution (LR) images. Dong et al. \cite{Dong2015BoostingOC} employ Convolutional Neural Networks (CNNs) for text image super-resolution (SR) and achieve significant outcomes in the ICDAR 2015 competition \cite{Peyrard2015ICDAR2015CO}. Nakao et al. \cite{Nakao2019SelectiveSF} develop a dual CNN framework comprising a character SR module and a general image SR module, trained on text images and ImageNet data, respectively. Quan et al. \cite{Quan2020CollaborativeDL} implement a multi-stage model strategy to precisely reconstruct the high-frequency details of LR text images. Reconstruction loss is also applied in various text image denoising tasks \cite{Lin2020BEDSRNetAD,Das2019DewarpNetSD, Dey2021LightweightDI, Feng2021DocTrDI, Georgiadis2023LPIOANetEH, Liu2023ShadowRO} to facilitate accurate background estimation.


\textbf{Coordinate Registration Methods}
Rather than concentrating solely on pixel fidelity, these methods aim to predict the transformation of coordinates in a distorted image to attain a more accurate and legible structure. This approach is especially beneficial for images in which text is warped due to perspective distortions or bending of the medium.

In the era of deep learning, document image dewarping has been modeled as a pixel prediction task. Das et al. \cite{Das2017TheCF} treat this task as one of semantic segmentation, utilizing a Fully Convolutional Network (FCN) \cite{Shelhamer2014FullyCN} to identify the visual characteristics of folds. Ma et al. \cite{Ma2018DocUNetDI} present a pioneering approach by defining the DID task as the process of determining the pixel displacement field from the distorted image, allowing for direct sampling of the distorted image to obtain the flatten image. They adopt a stacked U-Net \cite{Ronneberger2015UNetCN} for network and introduce a data synthesis method, enabling the automatic generation of large-scale document images and their corresponding displacement maps.

\textbf{Segmentation Methods}
Similarly, To eliminate noise in images, Hu et al. \cite{Hu2021RecycleNetAO} and Gholamian et al. \cite{Gholamian2023HandwrittenAP} approach the separation of overlapping text as a segmentation challenge using paired data. In tackling stamp-obscured images, Yang et al. \cite{Yang2023MaskGuidedSE} present a specialized erasure model that predicts binary masks for localizing stamps.

\subsubsection{Generative Learning}
Generative learning strives to produce new data points consistent with the distribution of given training data, 
paving the way for many image manipulation methods. These techniques primarily utilize Generative Adversarial networks (GANs) or diffusion models.

\textbf{Generative Adversarial Networks} Generative Adversarial Networks (GANs) \cite{mirza2014conditional,goodfellow2014generative} are the commonly used frameworks in the field of generation. GANs comprise two models: a generator and a discriminator. The generator aims to capture the distribution of authentic examples for generating new data instances. Conversely, the discriminator, typically a binary classifier, strives to distinguish between generated and real examples as accurately as possible. Through a minimax optimization process, the generator is believed to effectively capture the real data distribution.

Certain visual text processing techniques, like scene text removal\cite{zhang2019ensnet,liu2020erasenet}, can be regarded as image-to-image translation tasks where vanilla GANs are employed to learn the mapping from the input image to the output image, which solve the  following optimization problem: 
\begin{gather}
    L_{adv} = \mathbb{E}_x[\log D(x,G(x))] \\
    L_D = \mathbb{E}_{x,y}[\log D(x,y)] + \mathbb{E}_x[(1-\log D(x,G(x)))] \notag
\end{gather}
where $x$ is the given scene text image and $y$ is the ground truth. G and D are the generator and discriminator.

To achieve fine-grained control, conditional Generative Adversarial Networks (cGANs) \cite{chrysos2018robust} are commonly employed in manipulation tasks, where both the discriminator and generator are conditioned on additional information. For example, in scene text removal methods \cite{tursun2019mtrnet,tursun2020mtrnet++}, the text mask is provided to both the generator and discriminator as an extra input, accelerating the training convergence. To harness the potential of generative learning, more advanced architectures have been explored by Lyu et al. \cite{lyu2023fetnet} and Peng et al. \cite{peng2023viteraser}, along with the application of various pre-training techniques \cite{xie2022simmim}. Additionally, the background image imposes conditional constraints on style in scene text editing tasks \cite{wu2019editing}. Extending this framework, Kumar B G et al. \cite{subramanian2021strive} explore video scene text replacement with their innovative model, STRIVE, which includes target frame selection, modification, insertion, and critical restoration of video frames.

Extending the capabilities of cGANs, Cycle-consistent Generative Adversarial Networks (CycleGANs) \cite{zhu2017unpaired} resolve the issue of reference inapplicability for unpaired data. Utilizing CycleGANs, scene text can be synthesized \cite{wang1707adversarial,zhan2019spatial} by emulating the appearance space of real data. Moreover, a unified architecture \cite{Gangeh2021EndtoEndUD} is proposed by 
integrating deep mixture of experts with CycleGANs as the base network.

Karra et al. propose a StyleGAN \cite{Karras2018ASG} architecture, enabling the isolation of various factors, including hair, age, and sex, that influence the appearance of the final image, allowing for independent control over each. StyleGAN has been employed in \cite{krishnan2023textstylebrush} to extract an opaque latent style representation, disentangling the style and content of the source text image.


\textbf{Diffusion Models} Recently, diffusion models \cite{ho2020denoising,dhariwal2021diffusion,nichol2021glide} have demonstrated remarkable success in text-to-image tasks, offering an alternative and competitive solution to scene text manipulation. Unlike GAN-based approaches, diffusion models are not prone to training instability and mode collapse. Notably, recent advancements in diffusion models enable the incorporation of multi-modal conditional inputs. In particular, the Latent Diffusion Models (LDMs) \cite{rombach2022high} and similar works \cite{controlnet,nichol2021glide} offer readily available pretrained models capable of generating images from text prompts or image references.

Since most of visual text processing tasks can be modeled as a conditional generative paradigm, LDMs are used which can be categorized into two types:
synthesis-based and inpainting-based, distinguished by their respective conditions. The synthesis-based framework employs text prompts as conditions. For instance, text image super-resolution methods \cite{noguchi2024scene,zhao2023stirer} typically use texts from low-resolution images as input, whereas scene text editing \cite{wang2023letter,ji2023improving,santoso2023manipulating} specifies the target text as the condition. Conversely, inpainting-based methods often take an image with masked regions as the condition, incorporating a background style reference for scene text generation \cite{chen2023textdiffuser,ma2023glyphdraw,yang2023glyphcontrol}.

The loss function in LDM is defined as:

\begin{equation}
L= \mathbb{E}_{\varepsilon(x),y,\epsilon \sim \mathcal{N}(0,1),t} \left[ \left\| \epsilon - \epsilon_\theta (z_t, t, \tau_\theta(y)) \right\|_2^2 \right]
\end{equation}

For a given image $x$ and conditions $y$, the aim is to minimize the L2 norm of the difference between the actual noise $\epsilon$ and the predicted noise $\epsilon_\theta$, based on the noised latent representation $z_t$, the timestep $t$, and the conditioned representation $\tau_\theta(y)$.

\subsection{Related Research Areas}

This section provides a concise overview of scene text segmentation and editing detection, both of which are integral components closely related to the broader field of visual text processing methods.

\subsubsection{Scene Text Segmentation}
Scene Text Segmentation (STS) focuses on predicting pixel-level masks of text within an image, yielding a more detailed detection compared to the conventional bounding box prediction used in scene text detection. The outcomes of STS are crucial for tasks like scene text removal and editing, as they provide text stroke features essential for precise text localization.

Qin et al. \cite{qin2018robust} introduce a FCN approach \cite{Shelhamer2014FullyCN} specifically designed for text stroke segmentation.
This method employs the FCN to generate an initial and coarse text mask, which is subsequently refined using a fully connected Conditional Random Field (CRF) model. To address the discrepancies between synthetic and real-world text images, Bonechi et al. \cite{bonechi2020weak} develop a framework that leverages bounding box annotations of real text images to create weak pixel-level supervisions. Wang et al. \cite{wang2021semi} propose a semi-supervised method that utilizes real-world data annotated with either polygon-level or pixel-level masks. Their network features a mutually reinforced dual-task architecture, consisting of a single encoder and two decoders.

Xu et al. \cite{xu2021rethinking} introduce TextSeg, a comprehensive text dataset with fine annotations, and a novel text segmentation method, TexRNet. This dataset includes 4,024 images, featuring both scene and poster texts. TexRNet advances current segmentation techniques by incorporating key feature pooling and an attention module, thereby outperforming previous methods. Ren et al. \cite{ren2022looking} present a novel architecture, the Attention and Recognition enhanced Multi-scale segmentation Network, consisting of three main components: the text segmentation module, dual perceptual decoder, and recognition enhanced module.

In summary, scene text segmentation, though a niche field, holds distinct importance owing to its emphasis on text strokes and characters. 


\subsubsection{Editing Detection} The task of text editing detection, also known as tampered text detection, plays a critical role in safeguarding sensitive information.

Wang et al. \cite{wang2022detecting} stand out as pioneers in the field of tampered scene text detection, moving beyond traditional scene text detection models. Their method utilizes a shared regression branch capable of identifying global semantic nuances, complemented by specialized segmentation branches to distinguish between tampered and genuine text. Additionally, they propose a unique branch focused on frequency information extraction, acknowledging that manipulations are often more apparent in the frequency spectrum than in spatial domain.

The detection of tampered text in document images poses a unique challenge compared with general text editing detection, largely due to the the subtle visual clues associated with tampering.
To tackle this issue, Qu et al. \cite{qu2023towards} introduce a new architecture that combines visual and frequency features. This system also includes a multi-view iterative decoder, specifically engineered to utilize scale information for accurately identifying signs of tampering.


\begin{table*}[t]
\caption{The illustration of different text features and their employment in visual text processing methods.}   

    \centering
    \vspace{-10pt}
    \begin{tabular}{|c|c|c|c|c|c|}
    \hline
    \rowcolor{orange!30}
    \cellcolor{orange!20}-- & \cellcolor{orange!20}Text Structure & \cellcolor{orange!20}Text Stroke & \cellcolor{orange!20}Text Semantics & \cellcolor{orange!20}Text Style & \cellcolor{orange!20}Text Spatial Context \\
    \hline
    \cellcolor{orange!20}Text Image Super-Resolution & \cellcolor{orange!10} & \cellcolor{orange!10}\textbf{\checkmark} & \cellcolor{orange!10}\textbf{\checkmark} & \cellcolor{orange!10} & \cellcolor{orange!10} \\
    \hline
    \cellcolor{orange!20}Document Image Dewarping & \cellcolor{orange!10}\textbf{\checkmark} & \cellcolor{orange!10} & \cellcolor{orange!10} & \cellcolor{orange!10} & \cellcolor{orange!10}\textbf{\checkmark} \\
    \hline
    \cellcolor{orange!20}Text Image Denoising & \cellcolor{orange!10} & \cellcolor{orange!10} & \cellcolor{orange!10} & \cellcolor{orange!10} & \cellcolor{orange!10}\textbf{\checkmark} \\
    \hline
    \cellcolor{orange!20}Scene Text Removal & \cellcolor{orange!10} & \cellcolor{orange!10}\textbf{\checkmark} & \cellcolor{orange!10} & \cellcolor{orange!10} & \cellcolor{orange!10} \\
    \hline
    \cellcolor{orange!20}Scene Text Editing & \cellcolor{orange!10}\textbf{\checkmark} & \cellcolor{orange!10}\textbf{\checkmark} & \cellcolor{orange!10}\textbf{\checkmark} & \cellcolor{orange!10}\textbf{\checkmark} & \cellcolor{orange!10} \\
    \hline
    \cellcolor{orange!20}Scene Text Generation & \cellcolor{orange!10}\textbf{\checkmark} & \cellcolor{orange!10}\textbf{\checkmark} & \cellcolor{orange!10}\textbf{\checkmark} & \cellcolor{orange!10}\textbf{\checkmark} & \cellcolor{orange!10}\textbf{\checkmark} \\
    \hline
    \end{tabular}
\label{relation}
\end{table*}

\section{Delving into Text features}

In this section, we explore various crucial text-related features, including structure, stroke, semantics, style, and spatial context, and their significant function integration in different tasks. Table \ref{relation} shows their relationships.

\label{section3}
\subsection{Text Structure Feature}
Text structure, encompassing the layout, shape and orientation of text,
indicates how texts are arranged on a poster or in the wild. The structure feature helps maintain layout consistency between pre-enhanced and enhanced images. Furthermore, structure can be viewed as a unique style that requires control in certain scene text manipulation methods. The most widely used methods to represent text structure involve the text center lines or text contour control points as detailed below.

\subsubsection{Prior Assumption in Document Image Dewarping} 
To overcome the limitations of traditional DID methods on specialized equipment, researchers have explored low-level features such as illumination effects\cite{Wada1997ShapeFS,Zhang2009AUF}, boundaries\cite{Cao2003ACS}, and text lines\cite{Liu2015RestoringCD,Meng2012MetricRO,Kim2015DocumentDV,Lu2006DocumentFT,Zhang2002StraighteningWT}, which can be deemed as structured priors. Illumination-based methods establish the relationship between image and reflected light intensities to estimate spatial parameters. Conversely, methods focusing on boundaries and text lines\cite{Gatos2007SegmentationBR,Guo2019AFP,Stamatopoulos2011GoalOrientedRO} aim to estimate deformation parameters, working under the premise that text lines should appear horizontal and straight post-rectification. 

With the advanced data-driven neural networks, researchers leverage the mentioned prior assumption for model design and supervision to accelerate training process and enhance reconstruction performance. Under the text lines constrain, Jiang et al. \cite{Jiang2022RevisitingDI} refine the image by resolving an optimization problem that incorporates grid regularization. Based on the boundary assumption, Ma et al. \cite{Ma2022LearningFD} and Zhang et al. \cite{Zhang2022MariorMR} first detect the document's edges to remove the background, facilitating a preliminary dewarping step. Furthermore, Xie et al. \cite{Xie2022DocumentDW} introduce an approach that extracts sparse control points using an encoder architecture, followed by generating a dense displacement map through interpolation. The control points, strategically placed on the distorted image, are quantifiable, allowing for a tailored number to match the complexity of different datasets and offering manageable computational times.

\subsubsection{Layout Learning in Scene Text Editing and Generation} Yang et al. \cite{yang2020swaptext} introduce SwapText, which leverages a Thin Plate Spline Interpolation Network (TPS) \cite{wang2022tpsnet} to learn the spatial style of original texts. This comprehensive network adopts a divide-and-conquer strategy, featuring three specialized sub-networks: the text swapping network, the background completion network, and the fusion network. In the realm of generation, TextDiffuser \cite{chen2023textdiffuser} is conceived. A significant enhancement in TextDiffuser is the integration of a Layout Transformer \cite{gupta2021layouttransformer}, capable of learning text positions and layouts, coupled with a character-aware loss function to stabilize the training of the Diffusion Model.

\subsection{Text Stroke Feature}

Text stroke, denoting the appearance of a character glyph, plays a significant role in various visual text processing tasks acting as the guidance or condition. In practice, text stroke information can be acquired through human annotations or text segmentation techniques and represented by a binary segmentation map. The explicit construction of text stroke effectively eases the difficulty of pattern learning and contributes to elaborate processing.

\subsubsection{Fine-grained Guidance in Scene Text Removal, Editing, and Super-resolution} 

\textbf{Removal} Predicting fine-grained stroke features is crucial for accurate text erasure, and it provides a significant advantage over more general text region segmentation methods. A key part of this approach involves preserving the natural texture of the background while efficiently using its information. Keserwani et al. \cite{keserwani2021text} address this challenge by introducing a symmetric line character representation to improve stroke feature prediction. A specialized mask loss is employed to direct the network in learning essential features. Lee et al. \cite{lee2022surprisingly} further investigate the text stroke features by extracting text stroke region and text stroke surrounding region with weakly supervised learning. They use a gated attention mechanism to adjust confidence levels across these regions, leading to more precise segmentation of text strokes. Liu et al. \cite{liu2022don} devise a low-level contextual guidance block to capture image structural details, alongside a high-level contextual guidance block focusing on semantic aspects of the latent feature space. Moreover, they incorporate a feature content modelling block to blend the immediate pixels around text areas with the broader background, thereby minimizing texture inconsistencies in complex settings.

Contrasting with the aforementioned methods, Qin et al. \cite{qin2018automatic} and Tang et al. \cite{tang2021stroke} apply pre-trained text detection models for segmenting text regions prior to text removal, facilitating a more straightforward extraction of text strokes. The stroke masks are used to assist in reconstructing the background. A notable distinction between their approaches is that Tang et al. \cite{tang2021stroke} implement a sequential process, whereas Qin et al. \cite{qin2018automatic} use a parallel decoding strategy to integrate stroke characteristics effectively during inpainting.

A progressive strategy to enhance text stroke features is proposed in PERT \cite{wang2021pert}, in which an erasing block — merging a text localization network with a background reconstruction network — is repeatedly implemented in sequence. Significantly, PERT alters only text regions, keeping non-text regions intact under the direction of the learned text stroke. Lyu et al. \cite{lyu2022psstrnet} introduce a mask update module that incrementally refines text segmentation maps, employing attention mechanisms guided by the output of the previous iteration. Bian et al. \cite{bian2022scene} propose a comprehensive four-stage model, beginning with region-level mask processing through a detection-then-inpainting network. This generates a stroke-level mask and an initial coarse result, further enhanced by a follow-up network using both masks. Concurrently, Du et al. \cite{du2023progressive} detail an intermediate self-supervision approach based on the similarity of text stroke masks from augmented image versions, demonstrating enhanced performance in real-world scenarios through pretraining on synthetic data.

\textbf{Editing} In terms of scene text editing, Das et al. \cite{das2023fast} introduce an end-to-end framework that focuses on creating target masks for style transition. This process begins with defining the mask for the target image, using the source image and its glyph mask as references. By employing detailed masks, they enable a seamless exchange of background and font styles. Furthermore, Qu et al. \cite{qu2023exploring} develop stroke guidance maps to specifically delineate areas for editing. Unlike implicit methods that modify all image pixels, these explicit directions help isolate background distractions and direct the network's attention to specific text editing rules.

\textbf{Super-resolution} Text stroke is also employed in text image super-resolution tasks. Chen et al. \cite{Chen2021TextGS} propose a novel strategy that deconstructs characters into constituent strokes, and use stroke-level attention maps generated by an auxiliary recognizer to guide the super-resolution process for finer recovery. These developments make significant progress in enhancing the accuracy and quality of text image super-resolution. Additionally, Ma et al. \cite{Ma2023ABF} introduce a real-world Chinese-English benchmark dataset, and develop an edge-aware learning method supervised by a text edge map. In response to the complexity of character structures, Li et al. \cite{Li2023LearningGS} utilize the StyleGAN \cite{Karras2018ASG} \cite{Karras2019AnalyzingAI} to capture a wide range of structural text variations, leveraging generative structure priors for accurate text image restoration.

\subsubsection{Glyph Condition in Scene Text Generation} Text stroke offers explicit glyph conditions in scene text synthesis or generation, seamlessly integrating with conditional diffusion models \cite{rombach2022high}. Ma et al. \cite{ma2023glyphdraw} present GlyphDraw, an innovative framework aimed at precise control over character generation, enhanced by additional information like text locations and glyph features. However, GlyphDraw's notable limitation is its restricted capacity for producing varied text arrangements, such as multi-row or densely packed texts, a constraint linked to the dataset's scope. Tailored for explicitly learning text glyph features via ControlNet \cite{controlnet}, Yang et al. \cite{yang2023glyphcontrol} have developed GlyphControl, which utilizes a ``locked copy" technique to maintain the stability of pretrained diffusion models. In support of extensive training, they also introduce LAION-Glyph, a comprehensive benchmark for visual text generation, establishing a new benchmark in this domain.

\subsection{Text Semantic Feature}
Text sequences contain more than just a series of characters, and it also carries substantial semantic information that can direct the restoration or alteration of text images. For example, text can offer robust supervision in image super-resolution tasks, ensuring that text contents are preserved throughout the process. Additionally, this semantic richness enables generation or editing models to produce legible texts. The learning of semantic features can be facilitated either through an auxiliary text recognition task or by an independently pre-trained module capable of capturing semantic nuances.

\subsubsection{Semantic Intergration in Text Image Super-resolution} In the field of text image super-resolution, the semantic feature plays a crucial role. Wang et al. \cite{Wang2020SceneTI} propose TSRN, a novel architecture that incorporates bidirectional long short-term memory with residual blocks. This design effectively captures both sequential and contextual information in text images, significantly enhancing the reconstruction quality. Building on the foundation laid by TSRN, subsequent research introduces additional prior information and auxiliary constraints to further refine super-resolution techniques for text images. Chen et al. \cite{Chen2021SceneTT} utilize a pre-trained transformer for extracting supervised content, such as character position and contextual information. Zhao et al. \cite{Zhao2021SceneTI} present a parallel contextual attention network, aiming to capture sequence-dependent features and enrich the reconstruction with more high-frequency details. Ma et al. \cite{Ma2021TextPG} integrate a character probability sequence as auxiliary information, employing multi-stage refinement to progressively enhance low-resolution images. Continuing this trend, Ma et al. \cite{Ma2022ATA} develop  a transformer-based module  to synchronize text priors with spatially-deformed text images, ensuring accurate feature alignment.

Focusing on integrating implicit sequence features from vision with explicit semantic features from language, Zhao et al. \cite{Zhao2022C3STISRST} introduce the C3-STISR, a triple clue-assisted network. This network leverages recognition, visual, and linguistic cues to enhance super-resolution. Following this approach, Huang et al. \cite{Huang2023TextIS} adapt the triple clue framework, replacing the character probability sequence with a more complex semantic text embedding prior obtained from a text encoder. In a similar vein, Zhu et al. \cite{Zhu2023ImprovingST} develop a dual prior modulation network. This network utilizes both a text mask and recognition results as priors, aiming to further improve the process.

\subsubsection{Semantic Supervision in Scene Text Editing and Generation} In addition to super-resolution, semantic features are also introduced in editing and generation tasks. As prior information, Wang et al. \cite{wang2023letter} utilize a  character embedding to facilitate their image-to-image translation.
Addressing the nuances of text embedding presentation, Liu et al. \cite{liu2022character} highlight the significant impact of overlooking character-level input features on the fidelity of visual texts. Their study demonstrates that a shift from character-blind input tokens to character-aware tokens markedly improves the spelling precision of visual text. Furthermore, it is noteworthy that many scene text manipulation methods \cite{lee2021rewritenet,krishnan2023textstylebrush,qu2023exploring,su2023scene,susladkar2023towards,yang2023self} implicitly learn semantic features, often guided by auxiliary recognition loss.

\subsection{Text Style Feature}

Text style features encompass a range of inherent attributes such as font type, color, size, and shape, playing a crucial role in style-relevant text processing methods. These styles can be either implicitly learned in a latent space using deep neural networks or explicitly defined through fixed attributes, and facilitate both the visual consistency in scene text editing and the vivid synthesis in scene text generation.

\subsubsection{Style Reserving in Scene Text Editing} In the SRNet framework developed by Wu et al. \cite{wu2019editing}, a text conversion module is employed to alter the text content of the source image to match the target text, while preserving the original text style. Subsequently, the altered text content and inpainted background are input into a fusion module to produce the final edited text images. Concurrently, Zhang et al. \cite{zhang2021scene} introduce a network architecture akin to SRNet, but differentiated by a shared-weight background generation sub-network. This innovative feature simplifies training by facilitating the creation of integrated images.

Focus on character-level manipulation, Roy et al. \cite{roy2020stefann} present STEFANN, a two-stage method tailored for partial scene text editing. This approach encompasses two sub-networks: FANnet and Colornet. 
FANnet functions by receiving a character region image and a specified target character code, subsequently producing an image of the target character that retains the style of the source font. On the other hand, Colornet uses the produced target image and the original character image to perform colorization on the character.

To compensate for the scarcity of fully annotated data, the self-supervised training scheme \cite{li2021dense} is utilized to acquire a robust understanding of text styles. Lee et al. \cite{lee2021rewritenet} introduce RewriteNet framework that extensively uses real-world images. Its fundamental mechanism involves encoding text images into separate content and style features through dual encoders, and then amalgamating these features in the decoder. In a departure from conventional text editing methods, Krishnan et al. \cite{krishnan2023textstylebrush} propose Text Style Brush (TSB), a self-supervised technique that eliminates the need for target style supervision. This method exploits a wealth of real-world data to bridge domain gaps. Diverging from traditional methods that divide text editing into discrete stages like style transfer and background reconstruction, TSB adopts a more integrated approach. Expanding on TSB's proficiency in extracting both foreground and background styles, Yang et al. \cite{yang2023self} innovate in the cross-language scene text editing task. Their model architecture skillfully separates the learning of text content and style, enhancing the precision and versatility in manipulating scene text.

Beyond mere text editing, Su et al. \cite{su2023scene} present a groundbreaking task termed scene style text editing, a method that allows users to alter not only the text content but also its style. To facilitate this distinctive task, the team developed a background inpainting module responsible for extracting background textures. Following this, they introduce a foreground style editing module that encodes style into a high-dimensional latent space. Within this space, each encoding code vector represents a unique text style attribute, including factors such as rotation angle, font type, and color. As a result, text style editing is made possible by manipulating the code vectors in this latent space.


With the advance of the diffusion model, text style can also be represented with language prompts, based on which Ji et al. \cite{ji2023improving} tackle the scene text editing through a dual-encoder architecture wherein a CLIP text Encoder is utilized for style control. Besides, other diffusion-based methods realize style preservation of text editing through inpainting \cite{chen2023diffute} or conditional synthesis \cite{susladkar2023towards,santoso2023manipulating}, which effectively overcomes the shortcomings of earlier methods. However, these methods suffer from limited generalization and show incompetence with unseen style font.

\subsubsection{Style Transferring in Scene Text Generation}   
GANs have demonstrated a strong capability for style transfer, leading to their application in scene text synthesis \cite{wang1707adversarial,zhan2019spatial,fang2019learning,fogel2020scrabblegan}. Zhan et al. \cite{zhan2019spatial} introduce a concept of synthesis fidelity in both geometry and appearance spaces through their SFGAN. This model combines a geometry synthesizer with an appearance synthesizer: the former functions as a spatial transformation network, integrating background images with foreground text to ensure text alignment with the background plane, while the latter uses a cycle structure  to facilitate transition between synthetic and real image domains. Meanwhile, Fang et al. \cite{fang2019learning} unveil STS-GAN, a completely learning-based approach with a dual-stage structure consisting of a character generator and a word generator. The character generator, fed with a character label and a latent vector, creates styled character images utilizing both conditional adversarial and style losses. The word generator then processes the combined word image, balancing noise reduction with the preservation of character structure through L1 and adversarial losses. Additionally, Fogel et al. \cite{fogel2020scrabblegan} introduce Scrabble-GAN, which employs a semi-supervised method to produce handwritten text images varied in style and vocabulary. This architecture features individual character generators, a style-controlling discriminator, and a text recognizer to ensure legibility.

\subsection{Text Spatial Context Feature}

Text feature includes not only inherent characteristics but also the spatial contexts that represent the relationship between texts and their surroundings. This includes, but is not limited to, background texture, contour, and depth. Spatial context features of text can serve as auxiliary priors, augmenting tasks like document image dewarping, text image denoising, and scene text generation. Techniques such as depth estimation and contour prediction are employed to acquire this spatial context information.

\subsubsection{3D Reconstruction in Document Image Dewarping} Background spatial information, a significant feature of text spatial contexts, is employed in document image dewarping as a prior for rectification. Traditional hand-crafted methods for  DID rely on 3D reconstruction techniques, typically involving a two-step process: estimating the 3D shape of a warped document page and then flattening it. Specialized equipments, such as structured-lighting systems \cite{Brown2001DocumentRU,Sun2005GeometricAP}, laser range scanners \cite{Chu2005AFA} \cite{Zhang2008AnIP}, and structured laser beams \cite{Meng2014ActiveFO}, are used to gather the necessary 3D data. For the flattening phase, various methods are developed to approximate the physical model of paper deformation. Brown et al. \cite{Brown2001DocumentRU,Brown2004ImageRO} introduce a particle-based mass-spring model. Pilu et al. \cite{Pilu2001UndoingPC} estimate deformations using applicable surfaces. Expanding on this, Brown et al. \cite{Brown2005ConformalDO} use conformal mapping to parameterize the document’s 3D surface to correct warped images. Zhang et al. \cite{Zhang2008AnIP} implement a distance-based penalty metric, and Meng et al. \cite{Meng2014ActiveFO} apply extensible surface interpolation. However, these methods face limited practicality for routine use due to the requirement for additional hardware.

In the deep learning era, 3D information is adopted as the intermediate supervision in network training, disentangling the problem in a physically-grounded manner. With the proposed Doc3D dataset which contains 3D coordinate maps, Das et al. \cite{Das2019DewarpNetSD} first regress the 3D shape on the input document image and later perform the texture mapping for the final result. Das et.al.\cite{Das2021EndtoendPU} and Feng et.al.\cite{Feng2022GeometricRL} further inherit the strategy with different architecture design, enhancing the representation learning of the spatial
attributes that bridge the distorted image and the rectified image.


\subsubsection{Background Estimation in Text Image Denosing} In the restoration of noised images, estimating the background is essential, with various works focusing on illumination correction. Lin et al. \cite{Lin2020BEDSRNetAD} introduce  a substantial synthetic dataset, and a dual-component neural network. This network comprises a background estimation network and a shadow removal network, trained adversarially. Subsequent studies \cite{Das2019DewarpNetSD, Dey2021LightweightDI, Feng2021DocTrDI, Georgiadis2023LPIOANetEH, Liu2023ShadowRO} augment illumination correction networks using various criteria such as perceptual loss, and architectures including Transformers. Wang et al. \cite{Wang2022UDocGANUD} devise a light-guided network that employs cycle consistency constraints for unpaired data. This network aims to solve document illumination issues without inducing unwanted color shifts. Building on these advancements, Zhang et al. \cite{Zhang2023DocumentIS} present RDD, a large-scale real document dataset, and use a stacked U-Net architecture \cite{Ronneberger2015UNetCN} enhanced with a background extraction module to better adapt to real-world images.

\subsubsection{Background Attributes Integration in Scene Text Generation} Spatial contexts are fully explored in most scene text generation methods in order to achieve a visually convincing and coherent integration of text within the scene. Jaderberg et al. \cite{jaderberg2014synthetic} introduce a synthetic text generation engine called MJSynth, designed to emulate the distribution of scene text images. The MJSynth  comprises several key modules, each contributing to the authenticity of the generated text. These modules include font rendering, border/shadow rendering, base coloring, projective distortion, natural data blending, and noise addition. In the font rendering module, text font and other properties are selected, and words are rendered with horizontal lines or random curves on the foreground image layer. Border or shadow effects for words are rendered on an optional image layer in the border/shadow rendering module. Border or shadow effects for words are rendered on an optional image layer in the border/shadow rendering module. The base coloring process involves filling three image layers – foreground, background, and border/shadow – with uniform colors sampled from clusters derived from natural images. The projective distortion module applies random transformations to the foreground and border/shadow layers, introducing variability. In the last two modules, randomly-sampled image crops from datasets like IC13 and SVT are blended into the image layers, adding realistic texture. Noise is also introduced to the final composition of the three image layers. Yim et al. \cite{yim2021synthtiger} introduce SynthTIGER, a synthetic text image generator that follows a pipeline similar to MJSynth but with some notable differences. SynthTIGER comprises five key procedures: text shape selection, text style selection, transformation, blending, and post-processing. A significant distinction from MJSynth is SynthTIGER's addition of noise text to the word box image, simulating the appearance of text regions cropped from scene images and enhancing the realism of the synthetic text. The performance of SynthTIGER in scene text recognition shows improvements, even with a smaller quantity of synthetic data.

Gupta et al. \cite{gupta2016synthetic} introduce SynthText, a synthetic engine that diverges from conventional word box generation methods. SynthText synthesizes scene text images through a distinct pipeline, benefiting both text detection and recognition tasks. Initially, it segments images into contiguous regions, while simultaneously obtaining a dense pixel-wise depth map derived from a CNN. Subsequent modules estimate the local surface orientation for these regions. SynthText then renders text, varied in font and color, onto the chosen region, adapting to the derived surface orientation. Empirical results show the advantages of leveraging such high-fidelity synthetic data. Building on this, Chen et al. \cite{chen2017text} use the SynthText pipeline to expand their dataset, specifically to diversify training data for detecting text on traffic informational signs.

Building upon SynthText`s framework, Zhan et al. \cite{zhan2018verisimilar} enhance synthetic image quality by incorporating semantic coherence, saliency guidance, and text appearance considerations. Semantic coherence involves selecting regions suitable for text blending based on classifying semantic segmentation results against a predefined list. Saliency guidance comes from a saliency map generated by a dedicated model. For text appearance, image patches from datasets like IC13 are used. HoG features of background regions, along with the mean and standard deviation of the Lab mode for text regions, pair together. These paired features guide the brightness and color of text blending. Subsequently, Zhang et al. \cite{zhang2019scene} streamline the process into two main modules: region detection and text embedding. The detection module estimates a semantic score map and blending contour to identify optimal text regions. Meanwhile, the embedding module uses geometric transformations and a GAN generator to achieve seamless integration of text into images.

Unlike the methods that overlay text on static 2D images, more complicated properties are utilized to render text within 3D scenes, including deformation, shadow, and occlusion. Liao et al. \cite{liao2020synthtext3d} pioneer SynthText3D, a method for virtual scene text image synthesis using thirty 3D scene models from Unreal Engine 4 \cite{qiu2016unrealcv}. This methodology encompasses four modules: camera anchor generation, text region generation, text generation, and 3D rendering. A key feature in the text region generation module is the direct extraction of surface normal maps and normal boundary maps, complemented by the stochastic binary search algorithm to identify available text regions on each surface. Additionally, 2D text boxes undergo geometry transformations to project onto 3D regions, integrating diverse illumination effects into the scene.  Long et al. \cite{long2020unrealtext} develop UnrealText, an approach emphasizing greater interaction with virtual scenes for improved diversity and realism. Initially, a random walk algorithm, supported by ray-casting, automatically captures images from varied viewpoints. Subsequently, lighting conditions are randomized to simulate different environments. In text region generation, the central point of the initial proposal is projected, and re-initialized squares—whose horizontal sides are orthogonal to the gravitational direction—are generated and expanded to define a more precise text area.

\section{Benchmark Datasets}
\label{section4}
The swift advancement of visual text processing tasks and algorithms has been paralleled by a significant growth in datasets for training and evaluation. This section offers an overview of the prominent datasets, with key characteristics summarization in Table \ref{dataall} and  comprehensive review in the following discussion.

\begin{table*}[h]
\renewcommand{\arraystretch}{1}
\centering
\caption{Statistics of representative visual text processing datasets, including train and test data size, main language, source (where ``Syn" denotes synthetic and "Real" indicates real-world), type (whether captured in a scene, document, or designed poster), scope (referring to original images or cropped regions), method (either human-annotated or model-generated). } 
\label{dataall}
\vspace{-10pt}
\resizebox{1\linewidth}{!}{
\begin{tabular}{ccccccccc}
\toprule
Task & Dataset      & Year & Size   & Language          & Source & Type           & Scope  & Method \\ \midrule
\multirow{1}{*}{Text Image Super-Resolution} & TextZoom \cite{Wang2020SceneTI}        & 2020 & 21,740 & English      & Real       & Scene          & Region  & Human         \\ \midrule
\multirow{5}{*}{Document Image Dewarping}& DocUNet\cite{Ma2018DocUNetDI}       & 2018  & 130           & English           & Real      & Document      & Whole  & Human         \\
                                        & DRIC\cite{Li2019DocumentRA}          & 2019  & 1,300         & English           & Syn      & Document      & Whole  & Human  \\
                                        & Doc3D\cite{Das2019DewarpNetSD}         & 2019  & 100,000       & English           & Syn      & Document      & Whole  & Human  \\
                                        & DIR300\cite{Feng2022GeometricRL}        & 2022  & 300           & English           & Real      & Document      & Whole  & Human  \\
                                        & WarpDoc\cite{Xue2022FourierDR}       & 2022  & 1,020         & Englsih           & Real      & Document      & Whole  & Human  \\ \midrule
\multirow{5}{*}{Scene Text Removal}      & SCUT-Syn \cite{zhang2019ensnet}        & 2019 & 8,800  & English      & Syn        & Scene          & Whole  & Model         \\
     & SCUT-EnsText \cite{liu2020erasenet} & 2020 & 3,562  & English           & Real   & Scene          & Whole  & Human  \\
                                         & Bian et al.  \cite{bian2022scene}     & 2022 & 12,120 & Multilingual & Real + Syn & Scene          & Whole  & Human + Model \\
     & PosterErase  \cite{jiang2022self} & 2022 & 60,400 & Chinese           & Real   & Design         & Whole  & Human  \\
     & Flickr-ST \cite{lyu2023fetnet}   & 2023 & 3,004  & English           & Real   & Scene          & Whole  & Human  \\ \midrule
\multirow{2}{*}{Scene Text Editing}      & SynthText-Based \cite{gupta2016synthetic} & 2019 & -      & English      & Syn        & Scene          & Region & Model         \\
 & Tamper \cite{qu2023exploring}     & 2023 & 159,725    & English           & Real + Syn   & Scene          & Region & Human  \\ \midrule
     
\multirow{2}{*}{Scene Text Generation}   & MARIO \cite{chen2023textdiffuser}            & 2023 & 10M    & English      & Real       & Scene + Design & Whole  & Model         \\
                                         & LARION-Glyph \cite{yang2023glyphcontrol}     & 2023 & 10M    & English      & Real       & Scene + Design & Whole  & Model         \\ \bottomrule
\end{tabular}}
\end{table*}

\subsection{Dataset for Text Image Super-resolution}
\textbf{TextZoom}  The early text image super-resolution methods \cite{Dong2015BoostingOC, Peyrard2015ICDAR2015CO, Wang2019TextSRCT, Ledig2016PhotoRealisticSI, Nakao2019SelectiveSF, Wang2019TextAttentionalCG, Quan2020CollaborativeDL, Mou2020PlugNetDA} predominantly rely on synthetically generated datasets. Typically, these datasets are created by Gaussian blurring with down-sampling high-resolution images, an
approach that poorly replicates the complex degradation processes encountered in real-world scenarios. As a result,
models trained on such data often fail to generalize to actual text images. To overcome this limitation, Wang et.al.\cite{Wang2020SceneTI} introduce the first real-world dataset TextZoom, which contains camera-captured LR-HR text image pairs with varying focal lengths collected from two general image super-resolution datasets: RealSR\cite{Cai2019TowardRS} and SRRAW\cite{Zhang2019ZoomTL}. TextZoom includes 17,367 training pairs and 4,373 testing pairs, with the latter divided into three subsets to represent different levels of blurriness, namely easy (1,619 samples), medium (1,411 samples) and hard (1,343 samples) levels. Text labels, the types of bounding boxes and the original focal lengths are also provided.

\subsection{Dataset for Document Image Dewarping}
\textbf{DocUNet} The DocUNet dataset is proposed in Ma et.al.\cite{Ma2018DocUNetDI} and utilized as the benchmark for comparison. 65 paper documents captured by mobile cameras are collected in two distorted shapes, resulting in 130 images in total, along with the corresponding flat-scanned images as ground truth. The documents include various types, such as receipts, letters, fliers, magazines, academic papers and books. In order to assure the diversity of distortion in the benchmark, both easy cases (e.g., the documents with only one crease or one fold) and hard cases (e.g., the documents with heavy wrinkles) are included. Note that the benchmark contains both the original photos and the tightly cropped ones, and the latter is normally used for evaluation.

\textbf{DRIC} 
Unlike the previous method \cite{Ma2018DocUNetDI} directly synthesizing distorted training images in 2D based on the assumption of locally rigid fabric, Li et al. \cite{Li2019DocumentRA} generate data in 3D space with different lighting and camera settings by a rendering engine. Specifically, various electronic document images are collected as flattened ground truth and projected to the pre-defined distorted surfaces including perspective, curve and fold.
Furthermore, the exposure and gamma correction are randomly adjusted to obtain the final rendered images. Ground truth flow is stored in RGB image format where R and G channels are the 2D texture coordinates, and B channel is a binary mask indicating whether each pixel belongs to the document.

\textbf{Doc3D}
With the same physically-grounded manner, Das et.al. \cite{Das2019DewarpNetSD} create the Doc3D dataset in a hybrid manner using both real document images and rendering software. They first capture the 3D mesh of deformed paper and then render the image with various textures in software. In total, the Doc3D dataset contains 100,000 photo-realistic images with rich annotations, including 3D coordinate maps, depth maps, normals, UV maps, and albedo maps.

\textbf{DIR300}
To involve a more complex background and various illumination conditions in the test set, Feng et.al.\cite{Feng2022GeometricRL} build the DIR300 dataset which contains 300 real document photos. Concretely, the images are taken with different cellphones in different scenes under several distortions, involving curve, fold, flat and heavily crumpled documents. The ground truth images are captured before the collection of the distorted images.

\textbf{WarpDoc}
Collected by Xue et.al.\cite{Xue2022FourierDR}, WarpDoc consists of 1,020 camera images of documents with various paper materials, layout and contents. These images are warped into six types of deformation, including perspective, fold, curve, random, rotation and incomplete page, for fine-grained document restoration methods evaluation.

\subsection{Dataset for Scene Text Removal}
\textbf{SCUT-Syn} For the purpose of scene text removal, the SCUT-Syn leverages text synthesis technology \cite{gupta2016synthetic} on background scene images to produce a total of 8,800 samples. This dataset is divided into a training set with 8,000 images and a test set comprising 800 images.

\textbf{SCUT-EnsText} To bridge the disparity between synthetic images and real-world data, Liu et al. \cite{liu2020erasenet} introduce the SCUT-EnsText. This dataset comprises 3,562 diverse images, sourced from public scene text spotting benchmarks \cite{wei2022textblock}. Each image in the dataset has been meticulously annotated to offer visually coherent erasure targets, with the assistance of Adobe Photoshop technology by human labor. 

\textbf{Data from Bian et al.} Bian et al. \cite{bian2022scene} curated an extensive real-world multilingual dataset encompassing 12,120 images, with 11,040 allocated for training and 1,080 for testing. The dataset is characterized by annotations such as text-free images, region masks, and text stroke masks. The text from the collected images was manually removed using the inpainting tools in Photoshop to provide text-free images as ground truth.

\textbf{PosterErase}  PosterErase \cite{jiang2022self} is collected from e-commerce platforms and predominantly features posters with Chinese text. This dataset comprises 60,000 training images and 400 test images. Accompanying each image are detailed annotations, which include bounding box information, text content, and text-free images meticulously processed by human experts.

\textbf{Flickr-ST} Yu et al. \cite{lyu2023fetnet} introduce a real-world dataset comprised of 3,004 images, with 2,204 designated for training and 800 for testing. Distinctive features of this dataset include exhaustive annotations such as images with text removed, pixel-level text masks, character instance segmentation labels, character category labels, and character-level bounding box labels.

\subsection{Dataset for Scene Text Editing}
\textbf{SynthText-Based Data} Drawing inspiration from text synthesis technology \cite{gupta2016synthetic}, synthetic data has been crafted to facilitate the training of scene text editing models. Specifically, various fonts, colors, and deformation parameters can be employed to produce stylized text, which is then superimposed onto a background image. This approach enables the acquisition of ground truth encompassing the background, foreground text, and text skeleton. It is important to note that the volume of data produced for training and evaluation varies based on the methodology employed. 

\textbf{Tamper} Qu et al. \cite{qu2023exploring} have generated 150k labeled images for supervised training, and an additional 2k paired images, constituting the Tamper-Syn2k benchmark for training scene text editing models. Beyond synthetic datasets, real-world data derived from established scene text detection and recognition benchmarks, such as ICDAR 2013 \cite{IC13}, ICDAR 2015 \cite{IC15}, SVT \cite{SVT}, SVTP \cite{SVTP}, IIIT \cite{IIIT5K}, MLT 2017 \cite{icdar2017}, MLT 2019 \cite{icdar2019}, CUTE 80 \cite{CUTE}, and COCO-Text \cite{cocotext}, have been leveraged for testing. From these datasets, they meticulously curated a total of 7,725 images, excluding those that are heavily distorted or illegible, to form the Tamper-Scene benchmark for testing.

\subsection{Data for Scene Text Generation}
The majority of scene text image generation methods do not require specifically annotated datasets for training, except for certain generative model based approaches that necessitate large-scale text-image paired data. For instance, the MARIO \cite{chen2023textdiffuser} and LAION-Glyph \cite{yang2023glyphcontrol} benchmarks have been introduced, respectively.

\textbf{MARIO} The MARIO-10M dataset comprises approximately 10 million high-quality and diverse image-text pairs sourced from varied data sources, including natural images, posters, and book covers. It features detailed OCR annotations for each image, encompassing text detection, recognition, and character-level segmentation. Specifically, tools such as DB \cite{db}, PARSeq \cite{bautista2022scene}, and U-Net \cite{Ronneberger2015UNetCN} are employed for detection, recognition, and segmentation, respectively. The total volume of MARIO-10M is 10,061,720, with 10,000,000 samples allocated for the training set and 61,720 samples designated for the testing set.

\textbf{LAION-Glyph} The development of the LAION-Glyph dataset involves a series of steps.  First, utilizing the LAION-5B, a large-scale benchmark dataset \cite{schuhmann2021laion} designed for conditional generative models, an aesthetic score prediction model is applied to filter out low-quality images. Then, text-rich images are identified through OCR tools, which also include relevant annotations such as text bounding boxes and transcripts. Following this, the BLIP-2 model \cite{li2023blip} is employed to generate detailed captions. For practical purposes, the LAION-Glyph dataset is divided into three subsets: LAION-Glyph-100K, LAION-Glyph-1M, and LAION-Glyph-10M, each created using a random distribution approach.

\section{Performance Comparison}
\label{section5}
In this section, we present a tabulated performance analysis of the approaches  discussed. For each domain, we have selected widely available datasets for benchmarking purposes. The performance metrics are primarily sourced from the original publications, with exceptions duly noted.

\subsection{Text Image Super-resolution}
\subsubsection{Evaluation Metrics} 
To assess the fidelity of the super-resolved images, benchmark evaluations commonly employ the peak signal-to-noise ratio (PSNR) and structural similarity index measure (SSIM) \cite{wang2004image} metrics. Additionally, to evaluate the performance of downstream tasks, the recognition accuracy of the text recognition task is quantitatively adopted.

\subsubsection{Performance Comparison}
The effectiveness of text image super-resolution methods is comprehensively presented in Table \ref{TextZoom}. Notably, the final row of the table indicates the recognition accuracy of ground truth high-resolution images. According to the table, TSEPG \cite{Huang2023TextIS} emerges as the leading method in text image super-resolution, excelling in both recognition accuracy (64.68\%) and image quality (22.25 in PSNR and 0.7978 in SSIM). However, when compared to the recognition accuracy of 86.6\% achieved by high-resolution images, which represents the theoretical upper bound, there remains considerable margin for advancement in future research within this domain.

\begin{table*}[]
\renewcommand{\arraystretch}{1}
		\centering
		 \caption{Text image super-resolution methods on TextZoom. Bold denotes the \textbf{best} result, and underline denotes the \underline{second-best} result, same as the following.}
		\label{TextZoom}
  \vspace{-10pt}
  \resizebox{\linewidth}{!}{
		\begin{tabular}{ccccccccc}
			\toprule
			\multirow{2}{*}{Methods}                & \multicolumn{4}{c}{Recognition Accuracy}                         & \multicolumn{4}{c}{Image Quality (PSNR/SSIM)} \\ 
			\cmidrule(r){2-5} \cmidrule(r){6-9}
			 & Easy    & Medium  & Hard   & \bf{Average} $\uparrow$   & Easy    & Medium  & Hard   & \bf{Average} $\uparrow$  \\ \midrule
LR                     &62.40\%    &42.70\%    &31.60\%    &46.58\%    &  - &  - & -  &-\\
\midrule
Bicubic                     &64.70\%    &42.40\%    &31.20\%    &47.20\%    &22.3500/0.7884   &18.9800/0.6254   &19.3900/0.6592   &20.3500/0.6961\\
TSRN (2020)\cite{Wang2020SceneTI}  &75.10\%    &56.30\%    &40.10\%	&58.30\%    &25.0700/0.8897	&18.8600/0.6676	&19.7100/0.7302	&21.4200/0.7690\\
TPGSR (2021)\cite{Ma2021TextPG}    &78.90\%    &62.70\%    &44.50\%	&62.80\%	&23.7300/0.8805	&18.6800/0.6738	&20.0600/0.7440	&20.9700/0.7719\\
TBSRN (2021)\cite{Chen2021SceneTT} &75.70\%	&59.90\%	&41.60\%	&60.10\%	&23.8200/0.8660	&19.1700/0.6533	&19.6800/0.7490	&20.9100/0.7603\\
PCAN (2021)\cite{Zhao2021SceneTI}  &77.50\%	&60.70\%	&43.10\%	&61.50\%	&24.5700/0.8830	&19.1400/0.6781	&20.2600/0.7475	&21.4900/0.7752\\
TG (2021)\cite{Chen2021TextGS}     &77.90\%	&60.20\%	&42.40\%	&61.30\%	&23.3400/0.8369	&\underline{19.6600/0.6499}	&19.9000/0.6986	&21.4000/0.7456\\
TATT (2022)\cite{Ma2022ATA}        &78.90\%	&63.40\%	&45.40\%	&63.60\%	&\underline{24.7200/0.9006}	&19.0200/0.6911	&\underline{20.3100/0.7703}	&\underline{21.5200/0.7930}\\
C3-STISR (2022)\cite{Zhao2022C3STISRST} &79.10\%	&63.30\%	&\underline{46.80\%}	&\underline{64.10\%}	&- &- &- &21.5100/0.7721\\
TSEPG (2023)\cite{Huang2023TextIS} &\textbf{79.60\%}	&\underline{63.90\%}	&\textbf{47.50\%}	&\textbf{64.68\%}	&\textbf{25.3600/0.9053}	&\textbf{20.2600/0.6931}	&\textbf{20.5800/0.7782}	&\textbf{22.2500/0.7978}\\
DPMN (2023)\cite{Zhu2023ImprovingST}&\underline{79.25\%}	&\textbf{64.07\%}	&45.20\%	&63.89\%	&- &- &-	&21.4900/0.7925\\
\midrule
HR  &94.20\%	&87.70\%	&76.20\%	&86.60\% & - & - & - & - \\

\bottomrule
		\end{tabular}}
	\end{table*}

\subsection{Document Image Dewarping}
\subsubsection{Evaluation Metrics}
For DID quantitative evaluation, two key metrics are utilized: Image Similarity and OCR Accuracy. \textbf{Image Similarity:} Multi-scale structural similarity (MS-SSIM) \cite{1292216} employs SSIM across multiple scales via a Gaussian pyramid, assessing the global similarity between reconstructed and ground truth images. Local distortion (LD) \cite{4270276} calculates a dense SIFT flow \cite{Liu2011SIFTFD} from the reconstruction to the ground truth scan imames, gauging the rectification quality of local details. Ma et al. \cite{Ma2022LearningFD} further introduce aligned distortion (AD), which aligns the unwarped image and scan image before evaluating and weighs the error based on gradient magnitude, offering more robustness and accuracy against the limitations inherent in both MS-SSIM and LD. \textbf{OCR Accuracy:} Edit distance (ED) and character error rate (CER) \cite{1965Binary} are computed on selected text-rich images within the DocUNet to gauge the recognition quality of the reconstructions. However, due to variations in OCR engine selection and image datasets, OCR accuracy comparisons can be challenging and should be considered for reference only.

\subsubsection{Performance Comparison}


For performance benchmarking in the field of document image dewarping, we have selected DocUNet \cite{Ma2018DocUNetDI} and DIR300 \cite{Feng2022GeometricRL}, which are the two most extensively used datasets in this domain. Quantitative performance comparison for various DID methods is displayed in Table \ref{DocUnet}. Early methods like DewarpNet \cite{Das2019DewarpNetSD} focus on 3D construction and flatting, making it difficult to handle more detailed information. Later methods \cite{Feng2021DocTrDI}, \cite{Xie2022DocumentDW}, \cite{Feng2022GeometricRL}, \cite{Li2023LayoutAwareSD} begin to focus on global features (such as foreground images) and local features (such as text lines, control points, layout), achieving significant advancement. Notably, the most advanced solution to date, as proposed by \cite{Li2023LayoutAwareSD}, not only utilizes global and local feature, but also uses more realistic datasets compared to before, achieving a score of 0.526 in MS-SSIM and 6.72 in LD. This represents a significant improvement over earlier DID methods.

\begin{table*}[]
\renewcommand{\arraystretch}{1}
		\centering
		\caption{Document image dewarping  performance comparison on DocUNET and DIR300.  $\ast$ indicates experimental results  from the original paper of each method, with different OCR engine utilized. $\dagger$ indicates experimental results reported from \cite{gupta2021layouttransformer}, which use PyTesseract v0.3.9 for OCR testing.}
		\label{DocUnet}
  \vspace{-10pt}
  \resizebox{\linewidth}{!}{
		\begin{tabular}{ccccccccccc}
			\toprule
			\multirow{2}{*}{Methods}                & \multicolumn{5}{c}{DocUNet$\ast$}                         & \multicolumn{5}{c}{DIR300$\dagger$} \\ 
			\cmidrule(r){2-6} \cmidrule(r){7-11}
			& MS-SSIM $\uparrow$  & LD  $\downarrow$  & AD $\downarrow$    & ED $\downarrow$    & CER (\%)$\downarrow$  & MS-SSIM $\uparrow$  & LD  $\downarrow$  & AD $\downarrow$    & ED $\downarrow$    & CER (\%) $\downarrow$      \\ \midrule
			DocUNet (2018)\cite{Ma2018DocUNetDI}        &0.4100	&14.08        &-      &-                      &-          &-              &- &-&-&- \\   
			DewarpNet (2019)\cite{Das2019DewarpNetSD}   &0.4735	&8.95      &0.426  &1114.4 	      &26.92 &0.4921 &13.94 &0.331 &1059.57 &35.57  \\ 
   DFCN (2020)\cite{Xie2020DewarpingDI}        &0.4361	&8.50	      &0.434      &-                      &-          &0.5035 & 9.75 & 0.331 &1939.48 &50.99         \\
AGUN (2020)\cite{Liu2020GeometricRO}        &0.4491	&12.06        &-      &-                      &-          &-              &-  &-&-&-\\  
  Piece-Wise (2021)\cite{Das2021EndtoendPU}    &0.4879	&9.23         &0.468  &- &30.01 &-&-&-&-&- \\
   DWCP (2021)\cite{Xie2022DocumentDW}         &0.4769	&9.03	      &0.453      &-                      &-          &0.5524 &10.95 &0.357 &2084.97 &54.10            \\
   DocTr (2021)\cite{Feng2021DocTrDI}          &0.4970	&8.38	      &0.396 &\underline{576.4} &20.00 &0.6160 &7.21 &0.254 &699.63 &22.37 \\
   DocScanner (2021)\cite{Feng2021DocScannerRD}&\underline{0.5178}	&\underline{7.45}	      &\underline{0.334}	&632.3 &\textbf{16.48} &-&-&-&-&- \\
   PaperEdge (2022)\cite{Ma2022LearningFD}     &0.4700	&8.50	      &0.392 &1010.0 &22.10 &0.5836 &8.00 &0.255 &\textbf{508.73} &\underline{20.69} \\
   Marior (2022)\cite{Zhang2022MariorMR}       &0.4733	&8.08	      &0.403 &- &18.35 &-&-&-&-&- \\
   RDGR (2022)\cite{Jiang2022RevisitingDI} &0.4922	&9.36	  &0.461 &896.5 &20.68 &-&-&-&-&- \\
FDR (2022)\cite{Xue2022FourierDR}           &0.5000	&9.43	      &- &- &16.96 &-&-&-&-&- \\
DocGeoNet (2022)\cite{Feng2022GeometricRL}     &0.5040	&7.71 &0.380 &713.9 &18.21 &\underline{0.6380} &\underline{6.40} &\underline{0.242} &664.96 &21.89 \\ 
DocTr++ (2023)\cite{Feng2023DeepUD}         &0.5100	&7.52	      &-  &\textbf{447.5} &\underline{16.95} &- &-&-&-&-\\
Li et.al. (2023)\cite{Li2023LayoutAwareSD}  &\textbf{0.5260}	&\textbf{6.72}	&\textbf{0.300} &695.0 &17.50 & \textbf{0.6518} &\textbf{5.70} &\textbf{0.195} &\underline{511.13} &\textbf{18.91} \\

\bottomrule
		\end{tabular}}

	\end{table*}

\subsection{Scene Text Removal}

\subsubsection{Evaluation Metrics}
To evaluate scene text removal techniques, two evaluation protocols are employed: Detection-Eval \cite{nakamura2017scene} and Image-Eval \cite{zhang2019ensnet}. The Detection-Eval metric, developed by Nakamura et al. \cite{nakamura2017scene}, focuses on the thoroughness of text region removal. This metric uses an auxiliary text detector to gather detection results post text removal and evaluates the precision, recall, and F-score in accordance with the ICDAR-2013 \cite{IC13} and ICDAR-2015 \cite{IC15} standards, ensuring alignment with text localization ground truth. Often, the pretrained CRAFT \cite{baek2019character} is employed as the scene text detector, especially for assessing the performance of SCUT-EnsText, with a preference for the T-IoU measure \cite{liu2019tightness} to enhance result accuracy. In contrast, Image-Eval metrics defined in \cite{zhang2019ensnet}\cite{liu2020erasenet} emphasize the quality of the resulting images, including various metrics such as: (i) L2 error or mean squared error (MSE); (ii) PSNR, for peak signal-to-noise ratio comparison; (iii) SSIM \cite{wang2004image}, to measure structural similarity; (iv) AGE, which calculates the average of the graylevel absolute difference between the ground truth and the computed background image; (v) pEPs, which calculate the percentage of error pixels; and (vi) pCEPS, which calculates the percentage of clustered error pixels (number of pixels whose four-connected neighbors are also error pixels) Higher values of SSIM and PSNR, or lower values of AGE, pEPs, pCEPS, and MSE, indicate better performance. Additionally, Wang et al. \cite{wang2023real} have introduced BI-metric and EE-metric, focusing on background texture preservation and completeness of text erasure. For real-world datasets, only Detection-Eval methods are applicable, given the lack of ground truth background images. Human visual assessments also complement these evaluations on real datasets, providing a qualitative comparison of the effectiveness.
\subsubsection{Performance Comparison}
The effectiveness of various scene text removal methods is showcased on SCUT-EnsText and SCUT-Syn (Table \ref{enstext}), which are extensively employed for performance benchmarking in this domain. On SCUT-Syn dataset, MBE outperforms other methods in PSNR and SSIM thanks to its ensemble strategy.  In contrast, it can be seen than ViTEraser achieves the best performance in SCUT-EnsText in most of metrics. This is mainly because ViTEraser takes a self-training scheme for pre-training which learns more knowledge from real-world data.

\begin{table*}[]
\renewcommand{\arraystretch}{1}
		\centering
		\caption{Scene text removal performance comparison on SCUT-EnsText and SCUT-Syn.}
		\label{enstext}
  \vspace{-10pt}
  \resizebox{\linewidth}{!}{
		\begin{tabular}{ccccccccccc}
			\toprule
			\multirow{2}{*}{Methods}                & \multicolumn{7}{c}{SCUT-EnsText}                         & \multicolumn{3}{c}{SCUT-Syn} \\ 
			\cmidrule(r){2-8} \cmidrule(r){9-11}
			& PSNR $\uparrow$  & SSIM (\%) $\uparrow$  & MSE $\downarrow$    & AGE $\downarrow$    & pEPs $\downarrow$   & pCEPs $\downarrow$  & F $\downarrow$    & PSNR $\uparrow$    & SSIM (\%) $\uparrow$   & MSE $\downarrow$      \\ \midrule
			Pix2Pix (2017) \cite{isola2017image}          & 26.7000 & 88.56 & 0.0037 & 6.0860 & 0.0480 & 0.0227 & 47.0000 & 10.2000   & 91.08   & 0.0027   \\
			SceneTextEraser (2017) \cite{nakamura2017scene} & 25.4700 & 90.14 & 0.0047 & 6.0069 & 0.0533 & 0.0296 & 10.2000 & 25.4000   & 90.12   & 0.0065   \\ 
   EnsNet (2019) \cite{zhang2019ensnet}         & 29.5400 & 92.74     & 0.0024 & 4.1600 & 0.0307  & 0.0136 & 44.4000  & 37.3600 & 96.44     & 0.0021 \\  
   EraseNet (2020) \cite{liu2020erasenet}        & 32.3000 & 95.42     & 0.0015 & 3.0174 & 0.0160  & 0.0090 & 8.5000   & 38.3200 & 97.67     & \underline{0.0002} \\  
   Tang et al. (2021) \cite{tang2021stroke}  & 35.3400 & 96.24     & 0.0009 &- &- &- &- & 38.6000 & 97.55     & \underline{0.0002} \\
   Jiang et al. (2022) \cite{jiang2022self} &34.1400 & 89.15 &- &- &- &- &- &- & - &- \\
   CTRNet (2022) \cite{liu2022don} &\underline{35.8500} &\underline{97.40}  &0.0009 &- &- &-  &\underline{3.3000} &41.2800 &98.50 &{0.0002} \\
   PSSTRNet (2022) \cite{lyu2022psstrnet} &34.6500 &96.75 &0.0014 &\underline{1.7161} &0.0135 &0.0074 &9.3000 &39.2500 &98.15 &\underline{0.0002} \\
   MBE (2022) \cite{hou2022multi} &35.0300 &97.31 &- &2.0594 &0.01282 &0.0088 &- &\textbf{43.8500} &\textbf{98.64} &- \\
   SAEN (2023) \cite{du2023modeling}  &34.7500 &96.53 &\underline{0.0007} &1.9800 &0.0125 &0.0073 &-  &38.6300 &98.27 &0.0003 \\
   PERT (2023) \cite{wang2023real} &33.6200 &97.00 &0.0013 &2.1850 &0.0135 &0.0088 & 7.6000 &39.4000 &97.87 &\underline{0.0002} \\
PEN (2023) \cite{du2023progressive} &35.7200 &96.68 &\textbf{0.0005} &1.9500 &\underline{0.0071} &\textbf{0.0020} &3.9000 &38.8700 &97.83 &0.0003 \\
FetNet (2023) \cite{lyu2023fetnet}        & 34.6500 & 96.75     & 0.0014 & \underline{1.7161} & 0.0135  & 0.0074 & 10.5000  & 39.1400 & 97.97     & \underline{0.0002}  \\ 
ViTEraser (2023) \cite{peng2023viteraser}     & \textbf{37.1100} & \textbf{97.61}     & \textbf{0.0005} & \textbf{1.7000} & \textbf{0.0066} & \underline{0.0035} & \textbf{0.7680} & \underline{42.9700} & \underline{98.55}     & \textbf{0.000092} \\ \bottomrule
		\end{tabular}
  }
	\end{table*}

\subsection{Scene Text Editing}
\subsubsection{Evaluation Metrics}

Analogous to scene text removal, Image-Eval evaluation metrics such as MSE, PSNR, SSIM, and fréchet inception distance (FID) \cite{fid} are applied to synthetic datasets. Optimal performance is indicated by higher PSNR and SSIM values, alongside lower MSE and FID scores. In the context of real-world scene text images, the effectiveness of scene text editing is indirectly assessed by the accuracy of text recognition (SeqAcc).

\subsubsection{Performance Comparison }


It is critical to recognize that many scene text editing approaches predominantly utilize various synthetic datasets during their training and testing phases, leading to potential biases in performance evaluations. To promote a more balanced and fair assessment, our results are exclusively derived from the benchmark set by Qu et al. \cite{qu2023exploring}, as detailed in Table \ref{tamper}. Early methods like Pix2Pix \cite{isola2017image} focus on general style transfer and fail to dispose of fine-grained text images. With the divide-and-conquer model design, later methods \cite{wu2019editing,yang2020swaptext,qu2023exploring} significantly reduce the difficulty of pattern learning and achieve prominent improvement in both image visual quality and text rendering accuracy. Leveraging the advanced diffusion models, VTNet \cite{susladkar2023towards} demonstrates superior performance compared to other methods.

\begin{table}[h]
\renewcommand{\arraystretch}{1}
\centering
\caption{Scene text editing performance comparison on Tamper-Syn2k and Tamper-Scene.}
\label{tamper}
\vspace{-10pt}
\resizebox{1\linewidth}{!}{
\begin{tabular}{cccccc}
\toprule
\multirow{2}{*}{Methods} &
\multicolumn{4}{c}{Tamper-Syn2k}
&
\multicolumn{1}{c}{Tamper-Scene} \\
\cmidrule(r){2-5} \cmidrule(r){6-6}
                         & MSE $\downarrow$    & PSNR $\uparrow$  & SSIM $\uparrow$   & FID $\downarrow$   & SeqAcc $\uparrow$       \\ \midrule
Pix2Pix (2017) \cite{isola2017image}                  & 0.0732 & 12.0100 & 0.3492 & 164.2400 & 18.3820       \\
SRNet (2019) \cite{wu2019editing}                    & 0.0193 & 18.6600 & 0.6098 & 41.2600  & 32.2980       \\
SwapText (2020) \cite{yang2020swaptext}                & 0.0174 & 19.4300 & 0.6524 & 35.6200  & 60.6340       \\
MOSTEL (2023) \cite{qu2023exploring}                   & \underline{0.0123} & \underline{20.8100} & \underline{0.7209} & \underline{29.4800}  & \underline{76.7900}       \\
VTNet (2023) \cite{susladkar2023towards}                    & \textbf{0.0083} & \textbf{23.3600} & \textbf{0.7400}  & \textbf{26.2900}  & \textbf{79.8900}       \\ \bottomrule
\end{tabular}
}
\end{table}

\subsection{Scene Text Generation}
\subsubsection{Evaluation Metrics}

The primary goal of the scene text generation task is to facilitate large-scale pre-training for scene text detection and recognition models. The efficacy of this task is measured by the accuracy improvements in detection or recognition models following pre-training on the generated datasets. For techniques that create text bounding box images, enhanced recognition accuracy on text recognition datasets indicates a richer and clearer feature representation within the dataset with generated data. Conversely, for methods that generate whole scene text images, improved detection accuracy on text detection datasets is indicative of superior performance in text area distribution selection and realistic deformation rendering. Notably, some text-to-image works evaluate text rendering quality through various metrics. For example, the FID is used to compare the distribution of synthesized images against real images. Additionally, prompt fidelity is evaluated using OCR Evaluation and CLIPScore \cite{hessel2021clipscore}, with the latter measuring cosine similarity between image and text representations derived from CLIP \cite{radford2021learning}. Crucially, human evaluations also play an essential role in this  task. In these evaluations, participants are asked to rate the text rendering quality of generated images using structured questionnaires, providing a subjective yet vital perspective on the effectiveness of these methods.


\subsubsection{Performance Comparison }
In that the majority of scene text generation methods aim to assist text detection and recognition, we present experimental outcomes in Table \ref{gene1}, which details detection and recognition results using various synthetic datasets. In terms of the detection, VISD \cite{zhan2018verisimilar} has shown a significant improvement compared to SynthText \cite{gupta2016synthetic} due to its reasonable semantic selection mechanism. UnrealText \cite{long2020unrealtext} achieves the state-of-the-art results in all benchmarks because of its diverse text styles and complex backgrounds. When it comes to recognition, aside from UnrealText, SyntheTiger \cite{yim2021synthtiger} also showcases strong capacity, which aligns better with the real world data.

\begin{table*}[h]
\renewcommand{\arraystretch}{1}
\centering
\caption{Recognition and detection results  of ASTER \cite{ASTER}, BEST \cite{BEST} and EAST \cite{east}  trained on different synthetic data. $\dag$  means data copied from \cite{yim2021synthtiger}. }
\label{gene1}
\vspace{-10pt}
\resizebox{1\linewidth}{!}{
\begin{tabular}{cccccccccccccc}
\toprule
\multirow{2}{*}{Methods} &
\multicolumn{8}{c}{Recognition Results}
&
\multicolumn{5}{c}{Detection Results} \\
\cmidrule(r){2-9} \cmidrule(r){10-13}

                 & Recognizer        & Size    & IC13  & SVT    & IIIT5k   & IC15 &SVTP &CUTE80 & Detector  &Size &IC13 &IC15 &MLT17     \\ \midrule
MJSynth (2014) \cite{jaderberg2014synthetic}            &ASTER & 1M   & -    & \underline{39.2} & 51.6   & 35.7 & \underline{37.2} & 30.9 & - &- &- &- &-      \\
SynthText (2016) \cite{gupta2016synthetic}  &ASTER   & 1M   & -    & 30.3 & 53.5   & \underline{38.4} & 29.5 & \underline{31.2} & EAST   &10K &60.8 &46.3 &38.9     \\
VISD (2018) \cite{zhan2018verisimilar}       &ASTER  & 1M   & -    & 37.1 & \underline{53.9}   & 37.1 & 36.3 & 30.5    & EAST  &10K &74.8 &\underline{64.3} &\underline{51.4}     \\
SynthText3D (2020) \cite{liao2020synthtext3d} &- &-  &- &- &- &- &- &-  & EAST    & 10K  & \underline{75.6} & 63.4 & 48.3      \\
UnrealText (2020) \cite{long2020unrealtext} &ASTER  & 1M   & -    & \textbf{40.3} & \textbf{54.8}   & \textbf{39.1} & \textbf{39.6} & \textbf{31.6} & EAST  & 10K  & \textbf{78.3} & \textbf{65.2} & \textbf{54.2}    \\ \midrule
MJSynth (2014)\dag \cite{jaderberg2014synthetic}            &BEST    & 8.9M   & 83.5    & \underline{84.5} & 83.4   & \underline{66.0} & \underline{73.0} & \underline{64.6} & - &- &- &- &-      \\
SynthText (2016)\dag \cite{gupta2016synthetic}  & BEST      & 7M   & \underline{89.8}    & 82.5 & \underline{86.1}   & 64.5 & 69.1 & 60.1 & - &- &- &- &-   \\
SynthTIGER (2021)\dag \cite{yim2021synthtiger}   & BEST      & 10M   & \textbf{92.9}    & \textbf{87.3} & \textbf{93.2}   & \textbf{72.1} & \textbf{77.7} & \textbf{80.6}  & - &- &- &- &-  \\ \bottomrule
\end{tabular}}

\end{table*}

\section{Open Challenges}
\label{section6}
Notwithstanding recent advancements in visual text processing, numerous challenges persist. This section outlines key unresolved issues and potential future trends.

\subsection{Training Data}
The development of visual text image processing methods is significantly hampered by the scarcity of labeled real-world training data. For instance, acquiring paired source and target data with consistent source styles presents a notable challenge in scene text editing tasks. Additionally, training data obtained from the web are frequently contaminated with noise and subject to scene constraints. For example, benchmarks such as LAION \cite{schuhmann2021laion}, which are utilized for text image generation, predominantly comprise poster and web data, lacking in sufficient natural scene images.   Consequently, the compilation of comprehensive and high-quality datasets remains an unresolved issue in the field.

Future trends may pivot on optimizing the trade-off between quality and quantity. The question at hand is whether models perform better with weaker supervision across extensive datasets or with stronger supervision derived from smaller and high-quality datasets. In the former case, there is potential for the advancement of self-supervision or semi-supervision techniques to harness the potential of voluminous training data. For the latter, enhancing the model's generalizability in data-scarce situations through auxiliary approaches, such as domain adaptation, could be a valuable direction.

\subsection{Evaluation Metrics}
As previously mentioned, prevalent visual text processing techniques often rely on Image-Eval and Det/Rec-Eval metrics for assessment. However, the applicability of certain Image-Eval metrics, such as PSNR and SSIM, is limited due to the absence of ground truth pairs. Furthermore, general image and video quality metrics like FID may not be entirely appropriate for text image evaluations because of the domain shift from natural images to visual text images. Conversely, Det/Rec-Eval metrics can lead to skewed comparisons as various detectors or recognizers are employed across different methodologies. Additionally, the selection of hyperparameters and data augmentation techniques can significantly influence the outcomes.

A clear avenue for advancement in this field is the development of enhanced metrics tailored  for the text image domain. These metrics ought to be versatile and catering to a diverse array of text images that encompass multilingual types (such as English and Chinese), various shapes (including horizontal and oriented texts), and different environments (like posters and street scenes). Additionally, they should correlate closely with human judgment, facilitating accelerated and autonomous progress in methodological development with with minimal human intervention.

\subsection{Efficiency and Complexity}
 Efficiency remains a critical issue for visual text processing techniques. While many studies tout substantial accuracy gains, they often overlook reporting on model complexity (FLOPS) and inference speed (FPS). It is our contention that the majority of these methods have yet to find an optimal balance between accuracy and efficiency. This is largely due to the inherent architectural complexities, such as the self-attention mechanism in Transformers \cite{dosovitskiy2020image,vaswani2017attention} leading to intricate calculations, or the slow sampling rates in diffusion models \cite{ho2020denoising} that impede swift inference. Additionally, certain multi-stage approaches do not account for overall system efficiency, which limits their practical applicability. For example, methods for text removal should seamlessly incorporate a text detection mechanism to generate text masks.

A practical approach to enhance efficiency is the development of novel, streamlined architectures that reduce the time required for each denoising step in diffusion models \cite{li2023snapfusion} and decreasing computational complexity in Transformers. Techniques like model distillation also strive to improve efficiency. Furthermore, the use of end-to-end architectures can eliminate the need for auxiliary modules, streamlining the process further.

\subsection{Extension to Videos}
While 2D visual text image processing has advanced significantly due to technological progress and data availability, the evolution in higher-dimensional contexts, such as video, has been comparatively limited. The only video text processing method is STRIVE \cite{subramanian2021strive}, which aims for video scene text editing. The challenges in video-based visual text processing are manifold. Firstly, data availability and quality present substantial challenges. Although there is an abundance of raw video data, annotating this data to capture motion and temporal dependencies is a complex task. The lack of high-quality annotated data restricts the development of robust and generalizable models for processing visual text in videos. Secondly, the complexity of network architecture design poses another hurdle. Higher-dimensional data cannot be handled as simply as 2D images, which rely on discrete pixel values. Instead, they demand more sophisticated representations to manage long-range information crucial for interpreting temporal dynamics in videos and spatial relationships.

Future endeavors must focus on harnessing the plethora of online videos to curate high-quality video datasets, a task that requires substantial engineering and the development of dedicated automatic curation tools. Additionally, it is essential to craft video text processing architectures that are adept at managing high-dimensional data—akin to general video processing models—while also addressing the diverse attributes of text.

\subsection{Unified Framework}
Contemporary research in visual text processing often concentrates on frameworks designed for isolated tasks, neglecting the interconnected nature of these tasks. In practice, users typically present multifaceted needs. For example, within a single scene text image, a user might require concurrent operations such as removal, editing, and generation of text. Additionally, user interest often extends beyond textual elements to encompass various objects in the scene. A model with the ability to process text but without an understanding of the broader scene composition is considerably constrained.

Future research should focus on dismantling the barriers that segregate interrelated visual text processing tasks, with the goal of developing a cohesive and adaptable framework. Scene text erasure, for instance, could be treated as an intermediate step within a broader scene text manipulation procedure. In parallel, the development of a generalized image processing framework is warranted. Peng et al. \cite{peng2023upocr}  propose a unified framework to solve removal, segmentation and tamper detection. However, more tasks should be considered in the framework, which would be adept at simultaneously enhancing, altering, and synthesizing both text and common objects within images. To achieve this goal, it is imperative to refine tuning methods \cite{hu2021lora} for large-scale models, ensuring that enhancements in text processing do not compromise their core capabilities.

\subsection{User-friendly Interaction}
Current visual text processing approaches typically address all text regions in an image. However, users often need to tailor modifications to their individual requirements. To date, few studies in the fields of text removal \cite{mitani2023selective} and editing \cite{chen2023diffute} employ conditional models or extensive language models to facilitate precise content and style transfer. Despite this, such research is in its initial phase. The method of integrating diverse prompts or inputs for customized processing in various tasks presents a significant research opportunity.

The emergence of integrated visual models \cite{kirillov2023segment} and multimodal language-vision frameworks \cite{mao2023gpteval} has enabled the processing of diverse textual and visual prompts. Models such as SAM \cite{kirillov2023segment} now support visual prompts like points or bounding boxes to identify areas of interest, and advanced language models can interpret user-provided natural language instructions to derive precise image processing commands. Additionally, methods like in-context learning \cite{dong2022survey} and instruction tuning \cite{liu2023visual} are instrumental in translating personalized user instructions into specific visual text image processing results.

\section{Conclusion}
\label{section7}
In this paper, we comprehensively review recent progress in visual text processing tasks, which the first specialized survey to our best knowledge. We provide a hierarchical taxonomy that encompasses areas ranging from image enhancement and restoration to image manipulation, followed by specific learning paradigms. Additionally, we delve into text features closely related to the mainstream methods, including structure, stroke, semantics, style, and spatial context. Furthermore, we summarize the datasets for benchmarking and tabulate and compare the performance of existing approaches in various visual text processing tasks. Lastly, we share our perspective on the challenges and future directions for visual text image processing.


%



\ifCLASSOPTIONcompsoc
  \section*{Acknowledgments}
\else
  \section*{Acknowledgment}
\fi

This work is supported by the National Natural Science Foundation of China (Grant NO 62376266), and by the Key Research Program of Frontier Sciences, CAS (Grant NO ZDBS-LY-7024).

\ifCLASSOPTIONcaptionsoff
  \newpage
\fi



%



\bibliographystyle{IEEEtran}  
\bibliography{reference}  

\end{document}